\documentclass[10pt,twocolumn,letterpaper]{article}

\usepackage[pagenumbers]{cvpr} %

\usepackage{cuted}     %
\usepackage{caption}   %
\usepackage{booktabs}
\usepackage{multirow}
\usepackage{xspace}
\usepackage{amsmath,amssymb}
\usepackage{tikz}
\usetikzlibrary{positioning,arrows.meta,fit,backgrounds}

\newcommand{\method}{CoBoVLA\xspace}
\newcommand{\para}{LIBERO-Para\xspace}
\newcommand{\goal}{LIBERO-Goal\xspace}
\newcommand{\pizerofive}{$\pi_{0.5}$\xspace}

\definecolor{cvprblue}{rgb}{0.21,0.49,0.74}
\usepackage[pagebackref,breaklinks,colorlinks,allcolors=cvprblue]{hyperref}

\def\confName{}
\def\confYear{}

\title{CoBoVLA: Decoupling Language from Action with Discrete Codes\\for Robust and Interpretable Robot Manipulation}

\author{Mehul Arora$^{1}$\qquad Dhruv Agarwal$^{1}$\qquad Ambuj Agrawal$^{2}$\\[0.4em]
$^{1}$University of California San Diego\qquad $^{2}$Carnegie Mellon University}

\begin{document}
\maketitle
\begin{strip}
\centering
\includegraphics[width=\textwidth]{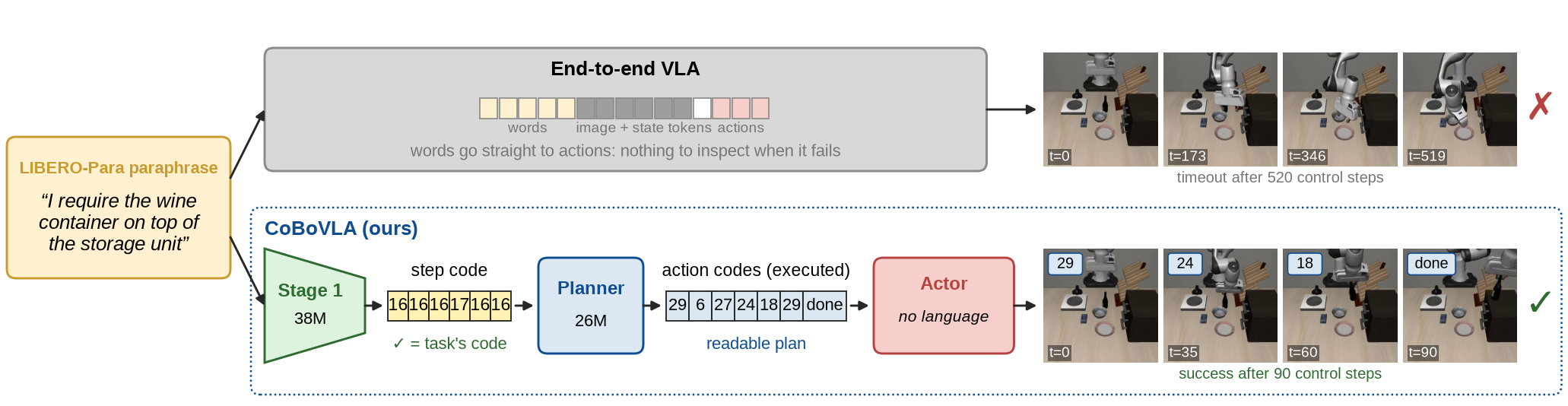}
\captionof{figure}{\textbf{The same paraphrase and initial state, end-to-end versus through discrete codes.}
Top: an end-to-end VLA maps the words directly to actions; it fails, and nothing in the model shows why (frames from SmolVLA~\cite{smolvla}).
Bottom: \method's stage~1 maps the paraphrase to the task's own step code, the planner turns it into a chain of action codes, and an actor executes it, and the task succeeds after 90 control steps ($t$).}
\label{fig:teaser}
\vspace{0.3em}
\end{strip}

\begin{abstract}
Vision-Language-Action (VLA) models achieve high success rates on the LIBERO benchmark but remain sensitive to paraphrasing.
Existing VLAs ranging from 0.6B to 7.5B parameters experience a 22--52 point performance drop on \para while SmolVLA exhibits an even larger decline, from 83\% to 8\%.
Furthermore, because most VLAs jointly learn language understanding and action generation, it is difficult to determine whether a failure originates from misunderstanding an instruction or executing it incorrectly.
To address these challenges, we introduce \method, or CodeBookVLA, a 250M-parameter VLA that decouples language understanding, planning, and control through learned discrete interpretable codes.
A text encoder maps the instructions to shared step codes, which a visual planner then converts into action code sequences executed by a language-independent actor.
This design improves robustness to instruction paraphrasing while allowing failures to be traced to individual stages.
It also enables efficient real-time control on resource-constrained hardware by encoding instructions only once and using compact discrete representations to keep each component lightweight.
\method achieves 91.4\% success on \goal, compared with 82.8\% for SmolVLA, under a matched protocol, with 17\,ms per control step latency.
On \para, it achieves 81.1\% across 4,092 unseen paraphrases and 78.5\% on matched episodes, compared with 36.5\% for SmolVLA LoRA fine-tuned on paraphrases.
Without two-step demonstrations, \method also completes 48.8\% of 160 free-form two-step tasks.
Finally, success reaches 91\% when instructions are correctly encoded but falls to 27.2\% otherwise, identifying language understanding as the main source of remaining failures.
Together, these results demonstrate that decoupling language understanding and control can improve instruction robustness while making VLA failures easier to diagnose.
\end{abstract}

\section{Introduction}
\label{sec:intro}

Vision-language-action (VLA) models map camera images and a language instruction to robot actions~\cite{pi0,pi05,openvla,smolvla}, and on the LIBERO benchmark~\cite{libero} the strongest ones now reach 96--98\% success~\cite{pi05,openvlaoft,vlaadapter,turbovla}.
These scores say little about language understanding, since a 0.5M-parameter policy that uses a lookup table of 40 task identifiers in place of language gets 95.8\% on LIBERO~\cite{minerva}, while paraphrasing the instructions reduces the success of VLAs with 0.6B--7.5B parameters by 22--52 points~\cite{liberopara}.

We argue that part of this difficulty comes from the architecture.
In a VLA, language and control are learned jointly in one network, so the representation of the instruction is mixed with the representation of the motion.
This leads to three problems.
First, the robustness to rewording has to be learned from robot data, or it has to be added at test time.
Second, when the robot fails, the model does not tell us whether it misunderstood the instruction or executed the motion wrongly (\cref{fig:teaser}).
Third, the whole network, including its language model, runs at every control step, which makes real-time control difficult on the resource-constrained hardware of a robot.

\begin{figure*}[t]
\centering
\includegraphics[width=\textwidth]{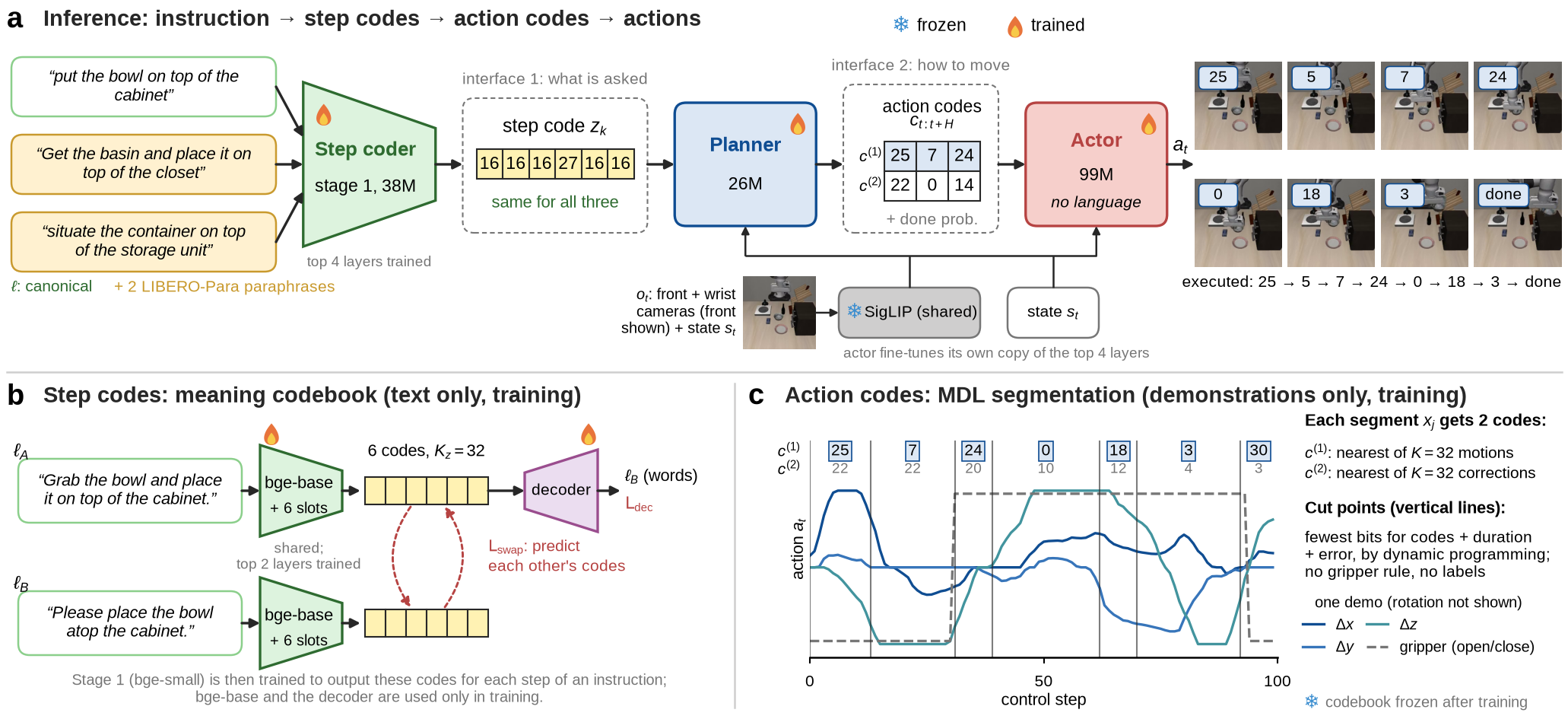}
\caption{\textbf{\method separates understanding from acting through two discrete codes.}
(a)~At test time, stage~1 maps the instruction to a learned step code; the canonical instruction and two paraphrases receive the same code.
The planner turns the current step code into a chain of action codes, and an actor turns the current action code into motor commands.
The frames show the first-level code being executed in a successful \para episode.
(b)~Step codes come from a meaning codebook trained on paraphrase pairs: the codes of each paraphrase must predict those of the other.
(c)~Action codes come from segmenting demonstrations by minimum description length.}
\label{fig:method}
\end{figure*}

To address these problems, we explore a different design, where understanding and acting communicate only through a discrete interface which can be inspected.
The proposed \method (\textbf{Co}de\textbf{Bo}ok VLA, \cref{fig:method}) has three stages.
A text encoder (stage~1) maps the instruction to a short sequence of discrete step codes, one for each step of the task, and it is trained so that paraphrases of the same instruction are mapped to the same codes (\cref{sec:method:stage1}).
Then, a visual planner takes the current step code together with camera images and robot state, and predicts a chain of action codes, \ie, discrete motion segments which are learned from demonstrations with minimum-description-length (MDL) segmentation (\cref{sec:method:codes,sec:method:planner}).
An actor which never sees the language converts the current action codes into motor commands (\cref{sec:method:actor}).
With this design, paraphrases that are mapped to the same step codes produce the same behavior, because no part after stage~1 sees the words.
Also, every failure can be localized to one stage by reading the codes, and the language stage can be improved with text only: all of the paraphrase robustness in this paper comes from training stage~1 on generated paraphrases, without collecting new robot demonstrations.
Finally, the architecture is chosen to run in real time on such hardware.
The instruction is encoded once, by a 38M-parameter text encoder, instead of by a language model at every control step, and each stage can be small because the discrete interface absorbs its errors: an imprecise output of a small model still selects the right code as long as it lands closer to that code than to any other.
On an RTX 4050 laptop GPU, one call of the planner and actor takes 74\,ms and is made every 5 control steps (17\,ms per step), well within the 250\,ms that these steps take at LIBERO's 20\,Hz control rate, while larger VLAs take 87--219\,ms per call on a faster RTX 4090~\cite{turbovla} (\cref{sec:exp:main}).

We evaluate \method on \goal~\cite{libero} and on all 4,092 paraphrases of \para~\cite{liberopara}, and compare it with SmolVLA~\cite{smolvla} and with SmolVLA fine-tuned on the same paraphrases (\cref{sec:exp:setup}).
With only 250M parameters, \method succeeds in 91.4\% of 500 standard \goal episodes (SmolVLA: 82.8\%, $p<10^{-4}$, two-sided sign test on matched episodes; all $p$-values in this paper use this test).
On \para, it succeeds in 81.1\% of the episodes without seeing any of the paraphrases during training, which is the highest success reported on this benchmark so far (under our protocol, \cref{tab:reported}).

Moreover, the step codes allow \method to follow one sentence which asks for two tasks in sequence.
Without any two-step demonstration, \method completes both steps in 48.8\% of 160 episodes, while SmolVLA does not complete the first step in any of the 80 episodes (\cref{tab:twostep}).

Analyzing the codes also shows where the remaining errors are (\cref{sec:exp:attribution}).
When stage~1 outputs the canonical step codes, the success is 91.0\%, which is close to the success with standard instructions (91.4\%), but when it does not, the success drops to 27.2\% (\cref{tab:attribution}).

\noindent In summary, our contributions are as follows:
\begin{itemize}
\item We propose \method, a VLA which connects language understanding and control only through discrete step codes and learned action codes, designed to run in real time on a laptop GPU (\cref{sec:method}). Replacing the step codes with their continuous pre-quantization vectors lowers the success on both benchmarks (\cref{sec:exp:dev}).
\item We provide a controlled evaluation on \goal and the full \para benchmark under one protocol, with a one-time confirmation on held-out initial states (\cref{sec:exp}).
\item We show the execution of free-form two-step instructions and attribute the failures to parsing, step switching and execution (\cref{sec:exp:twostep}).
\item We present a code-level failure analysis on \para which separates understanding, planning and execution errors, and places most of the remaining paraphrase gap in the language stage (\cref{sec:exp:attribution}).
\end{itemize}

\section{Related Work}
\label{sec:related}

\paragraph{Vision-language-action models.}
Most VLAs fine-tune a pretrained vision-language model to output robot actions, as discrete tokens in OpenVLA~\cite{openvla} or through a continuous action head trained with diffusion~\cite{diffusionpolicy} or flow matching~\cite{pi0,pi05,openvlaoft,xvla}.
Several works make them smaller and faster: SmolVLA~\cite{smolvla} and VLA-Adapter~\cite{vlaadapter} build on backbones of about 0.5B parameters, NanoVLA~\cite{nanovla} routes between a small and a large model, and TurboVLA~\cite{turbovla} drops the language model from the control path and gets 97.7\% on LIBERO with 0.2B parameters.

\paragraph{Robustness to instruction wording.}
Several benchmarks have shown that VLAs are sensitive to the phrasing of instructions~\cite{liberoplus,langgap,liberopara,liberopro}.
Data augmentation trains the policy on rephrased instructions~\cite{relabel,paraconsensus,synthaug,dial}; on LIBERO-Spatial with their own paraphrases, Li \etal~\cite{paraconsensus} remove nearly all of the paraphrase cost by LoRA fine-tuning of OpenVLA on paraphrases.
RoVLA~\cite{rovla} adds consistency constraints across rephrasings.
At test time, Watts and Cui~\cite{rephrase} have an LLM rewrite the instruction, which takes seconds per instruction, and Kwok \etal~\cite{cover} verify candidate actions against rephrasings at the price of many policy calls per step.
The work closest to ours is GSR/ParaVLA~\cite{paravla}, which decouples task semantics from visual features and reaches 72.5\% on \para with a 0.33B model trained on canonical instructions only.
In all of these methods, the instruction is still an input to the policy; in \method only stage~1 reads it, so the controller is invariant to wording and each failure can be traced to a stage.

\paragraph{Discrete skills and latent plans.}
LISA~\cite{lisa} maps language to discrete skill codes that condition a language-free policy and studies their interpretability, and LADS~\cite{lads} trains a language-regularized discrete latent-plan space and a high-level policy over it in separate stages.
SkillDiffuser~\cite{skilldiffuser} plans over interpretable skill abstractions.
QueST~\cite{quest} and STAR~\cite{star} learn discrete skill tokens on LIBERO, and several recent VLAs plan in a coarse discrete space before acting~\cite{dasl,libravla}.
Our action codes come from an MDL segmentation of the demonstrations into variable-length segments, following CompILE~\cite{compile}, LOVE~\cite{love} and LAST~\cite{last}.
These build on the minimum-description-length principle~\cite{rissanen}; PRISE~\cite{prise} and Subwords as Skills~\cite{subwords} instead compress quantized action sequences with byte-pair encoding, and earlier work segmented demonstrations by changepoint detection~\cite{cst}.
What we add is the use of such an interface for robustness to wording in a VLA, with a language stage trained on paraphrases alone, and the use of the codes to attribute failures on a paraphrase benchmark.

\section{Method}
\label{sec:method}

\method maps an instruction $\ell$ and observations $o_t$ to actions $a_t$ through two discrete interfaces (\cref{fig:method}), where $t$ indexes control steps.
The observation $o_t$ consists of two camera images (front and wrist) and the robot state $s_t\in\mathbb{R}^8$ (end-effector position and orientation and the two gripper-finger positions), and the action $a_t\in\mathbb{R}^7$ is a change of end-effector position and orientation and a gripper command.
Stage~1 maps $\ell$ to a sequence of step codes $z_{1:S}$, where $S$ is the number of steps in the instruction and each $z_k$ is a tuple of $M=6$ discrete codes (\cref{sec:method:stage1}).
For the current step $k$, the planner maps $(z_k, o_t)$ to a chain of action codes $c_{t:t+H}$: the codes of the motion segment that is executing at step $t$ and of the segments that follow it, $H=3$ segments in total (\cref{sec:method:planner}).
The actor maps $(c_{t:t+H}, o_t)$ to a chunk of actions and never receives $\ell$ or $z$ (\cref{sec:method:actor}).
We first describe the action codes (\cref{sec:method:codes}).
All visual inputs use the SigLIP~\cite{siglip} vision encoder of SmolVLA~\cite{smolvla} at $256\times256$ resolution, shared by the planner and the actor.

\subsection{Action codes from MDL segmentation}
\label{sec:method:codes}
We segment each demonstration into variable length motion segments.
This motion segment is represented using a two level code using a VQ-VAE style encoder~\cite{vqvae} with residual vector quantization~\cite{soundstream}.
The code is represented as $c=(c^{(1)},c^{(2)})$, where $c^{(1)}$ and $c^{(2)}$ are each chosen from its own learned codebook of $K=32$ entries.
The first level covers coarser motion covering plan level details while the next level covers more refined correction codes which level~1 missed.
We follow minimum description length (MDL) principle to come up with the best segmentation that describes each demonstration in the fewest bits.
Cutting too finely pays for many codes and cutting too coarsely pays in reconstruction error.
\begin{equation}
\begin{split}
\mathcal{L}(\text{seg}) = \sum_{j} \Big[ & -\log p(x_{j} \mid c_j) + \log K + \log D \\
& - \log p\big(c^{(1)}_j \mid c^{(1)}_{j-1}\big) \Big],
\end{split}
\label{eq:mdl}
\end{equation}
where $\mathcal{L}(\text{seg})$ is the description length of a segmentation of one demonstration into segments $j=1,2,\dots$, $x_j$ is the action trajectory of segment $j$, $c_j$ is its code, and $-\log p(x_j\mid c_j)$ is its reconstruction error under a Gaussian with re-estimated variance; $\log K$ and $\log D$ pay for the second-level code and the duration, where $D=187$ is the number of allowed segment lengths (6 to 192 control steps), and the first-level code is paid for under a per-task bigram prior, which makes code sequences that recur in a task's demonstrations cheaper.
We alternate dynamic-programming segmentation under \cref{eq:mdl} with re-training the residual autoencoder, starting from fixed 16-step windows, for 8 rounds~\cite{last,love,compile}, a hard-EM scheme as in segmental $k$-means~\cite{segkmeans}.
The segment granularity is determined by the data: initializations of 16, 32 and 64 steps all move toward the same length, and the converged segmentation places a boundary within three control steps of 87\% of gripper changes, although boundaries are never tied to gripper events.
The codebook is learned once on all 1,693 LIBERO demonstrations~\cite{libero} and then frozen.

\subsection{Stage 1: instructions to step codes}
\label{sec:method:stage1}
We use bge-base~\cite{bge}, top two layers fine-tuned, to encode the sentence into learned slots ($M=6$).
These slots are built by attending to the word tokens.
All slots share a learned cosine codebook of $K_z=32$ entries, and the slot values are replaced by their nearest codebook entries.
The training signal comes from the paraphrase pairs $(\ell_A,\ell_B)$ of the same instruction.
The codes of $\ell_A$ must predict the codes of $\ell_B$ and support decoding $\ell_B$.
We also add commitment loss, a batch-entropy term against collapse, and a penalty on repeated slots.
This is done to maintain the robustness across paraphrasing to allow the codes to learn the meaning rather than attaching words to them.

\paragraph{Paraphrase data.}
We generate 27,222 paraphrases with a large language model (Claude Sonnet 5.5) for the 40 LIBERO instructions and 191 recombined instructions built from the same objects and actions, and remove the 74 that exactly match a \para paraphrase.
\para is used only for testing.

\paragraph{Step coder.}
Instructions can describe more than one step.
The deployed stage~1 is a smaller step coder (bge-small~\cite{bge}, 38.4M parameters, top four layers fine-tuned) with one query embedding per step and a presence head that predicts how many steps the instruction contains.
It is trained to output, for every step, the meaning-codebook codes of that step's canonical instruction, on single-step paraphrases, on two-step sentences formed on the fly by joining paraphrases of two steps, and on about 1,500 two-step rewrites written by a language model.
Pairs that form one of the 16 two-step test pairs, in either order, are excluded from all of them.

\subsection{Planner: step codes to action-code chains}
\label{sec:method:planner}
The planner is a 6-layer transformer (width 512, 26.1M parameters).
At each call it predicts the next 3 action-code segments $c_{t:t+H}$ as a sequence of discrete tokens, plus a done probability for the current step.
It is an autoregressive decoder: it cross-attends to a memory of 128 frozen SigLIP tokens (each camera's $16\times16$ patch grid average-pooled to $8\times8$), one robot-state token and the embeddings of the six step codes, and greedily decodes two tokens per segment, one per level, followed by an end-of-chain token.
We train it with cross-entropy on the code chains of the \goal demonstrations, using the step codes of a randomly drawn paraphrase of the task.

In a closed loop, a planner trained with behavior cloning alone drifts out of the demonstration distribution, so we run one round of DAgger~\cite{dagger}.
We roll out the policy from initial states 10--29 (disjoint from the evaluation states 0--9), label each visited state with the code of the nearest training frame of the same task in terms of SigLIP similarity and state distance, and retrain on the demonstrations together with the relabeled states.
This nearest-frame lookup is only used to make training labels; at test time all codes come from the planner.
The done probability is the probability that the chain starts with an end-of-chain token.
We fine-tune it with binary cross-entropy on 80 two-step rollouts recorded with oracle step switching and labeled by whether the goal of the current step holds; these use the 16 two-step test pairs, but initial states 10--14 rather than the test initial states 0--9.

\subsection{Actor: action codes to motor commands}
\label{sec:method:actor}
The actor is a flow-matching transformer~\cite{flowmatching} (12 layers, width 512) that predicts a chunk of future actions from the SigLIP image tokens, the robot state and the current action codes.
It gets no language input.
Each of its blocks applies self-attention over one sequence of 128 pooled SigLIP tokens, a state token, six code-slot tokens (the two-level codes of the three planned segments) and 50 noisy action tokens, and the flow time and the code slots modulate every block through adaLN-zero.
Before entering this sequence, each code-slot token passes through two such blocks and then cross-attends to the full $16\times16$ patch grid of both cameras, so that the codes decide where the actor looks at full resolution.
We fine-tune the last four SigLIP layers at $0.2\times$ the learning rate, which together with the actor gives 99.4M trainable parameters.
An auxiliary loss supervises the attention from the code slots to the image tokens with the future path of the gripper in the image plane.
The actor is trained on \goal demonstrations with ground-truth codes and, with probability 0.5, with the codes the planner predicts for the same frame, so that it learns to cope with planner errors.

At test time, planner and actor are called every 5 control steps, and the predicted actions are executed with temporal ensembling~\cite{aloha}: an arm action is the $0.9^{\text{age}}$-weighted average of the predictions for that step from all chunks issued in the last 20 steps, where age is the number of control steps since the chunk was issued, and the gripper command comes from the newest chunk.

\subsection{Multi-step instructions}
\label{sec:method:multistep}
For an instruction with $S>1$ steps, the policy starts with $z_1$.
When the planner's chain starts with the done token in three consecutive calls, a short scripted motion moves the arm back to its starting pose (median 57 control steps), the ensembling buffer is cleared and the planner continues with $z_2$, as in SeqVLA~\cite{seqvla}.
No component is trained on two-step demonstrations; the sequence itself comes from stage~1.

\section{Experiments}
\label{sec:exp}

We perform four experiments.
(1)~We compare \method's performance against SmolVLA, a small VLA with 0.6B parameters, on both standard and paraphrased instructions under a single protocol (\cref{sec:exp:main}).
(2)~Evaluate \method on free-form two-step instructions without two-step demonstrations (\cref{sec:exp:twostep}).
(3)~Compare \method against existing literature on the \para (\cref{sec:exp:reported}).
(4)~Study what the learned codes reveal about the failures (\cref{sec:exp:attribution}).
The concluding sub-section, \cref{sec:exp:dev}, analyzes the design, and concludes why codes need to be discrete.

\subsection{Setup}
\label{sec:exp:setup}

On \goal~\cite{libero}, we run all 500 episodes, spread across 10 tasks with 50 different initial states in shared kitchen scenes.
Some baselines and ablations are run only on Goal-100, the first 10 initial states (0--9) of each \goal task (100 episodes); ParaVLA~\cite{paravla} uses the same 100 episodes for its \goal results and for its diagnostic and ablation experiments.
\para~\cite{liberopara} contains 4,092 paraphrases of the 10 \goal instructions, grouped by what they change: the action expression (870), the object reference (259), or both (compound, 2,963), and we run each paraphrase once.
Paraphrase $i$ of a task starts from initial state $i \bmod 50$.
For the most expensive baseline runs, we used a fixed 200-episode subset with about 20 paraphrases per task.
For the two-step instructions, we built 16 tasks from ordered pairs of \goal tasks whose goals could be held together; success was defined as both goals held at the end of the episode.
For even initial states, a sentence was written by a language model that fused two steps, for example: ``Pull the cabinet's middle drawer out and then move the wine bottle onto the rack'', while for the odd ones, two \para paraphrases were joined, for example: ``Position the serving bowl on the stove after you open the middle compartment of the cupboard''.
For each of the tasks, we ran from initial states 0--9, giving a total of 160 episodes.

\paragraph{Evaluation protocol.}
All results in \cref{tab:twostep,tab:twostepfail,tab:attribution} come from one simulator setup (MuJoCo with EGL rendering, $256\times256$ images, episodes of at most 520 steps for single-step and 980 for two-step tasks).
We found that the instruction given to the policy matters: some LIBERO forks read it from the language field of the task file, which rewords 5 of the 10 \goal tasks (\eg ``open the middle layer of the drawer'').
On these five tasks (initial states 0--9), SmolVLA succeeds in 11/50 episodes with the reworded field and 37/50 with the task-file names, and \method in 28/50 and 46/50.
We use the task-file names, as in standard LIBERO and in the training data.
Under this protocol, the public SmolVLA LIBERO checkpoint reaches 82.8\% on \goal, consistent with an independent rerun that gives 84.7$\pm$3.7\%~\cite{smolmamba}.

We defined two baselines, SmolVLA and SmolVLA + paraphrase LoRA.
For both, we used the 605M-parameter released checkpoint of the SmolVLA~\cite{smolvla} model.
To separately study the architecture from the paraphrase data, in SmolVLA + paraphrase LoRA, we fine-tune the SmolVLA with LoRA rank 32 on the language model and action expert (27.2M trainable parameters) for 3,000 steps, on \goal demonstrations whose instruction is replaced with a probability of 0.8 by a paraphrase from the same 27,222 paraphrases that train our stage~1.
Training and validation losses stop decreasing after about 2,000 steps.

Design choices were made on initial states 0--9 and 30--39 of \goal only.
\para was never used for selection.
After freezing the configuration we evaluated it once on the untouched initial states 40--49 and obtained 89\%, against 94\% and 88\% on the two selection sets.
Initial states 10--29 were used to collect the DAgger data (\cref{sec:method:planner}), so we also report the 300 episodes that exclude them.

\subsection{Comparison under one protocol}
\label{sec:exp:main}

We compare our \method with SmolVLA and its paraphrase LoRA variant.
On the standard instructions, \method achieves a 91.4\% success rate across 500 episodes of \goal, whereas SmolVLA achieves 82.8\%.
The gap holds on the 300 episodes that excluded DAgger initial states (90.3\% vs.\ 84.0\%).
Particularly, SmolVLA is better than \method on two tasks, pushing the plate (45 vs.\ 40 out of 50) and the wine rack (42 vs.\ 39), and \method is better or equal on the remaining eight.

On the paraphrases, a much larger difference was observed.
SmolVLA succeeds in 8\% of the subset episodes, consistent with the reported 4.5\% on the full benchmark under another protocol~\cite{paravla}.
LoRA fine-tuning of SmolVLA on our paraphrases pushes performance to 36.5\% but costs 16 points on the standard instructions (83 $\to$ 67 on Goal-100).
\method achieved 78.5\% on the same 200 episodes and 81.1\% on all 4,092 while maintaining the success rate on the standard instructions.
The paraphrase drop is a modest 10.3 points.
On the two-step instructions, SmolVLA failed on all of the 80 episodes in the first step, while \method completed both steps in 48.8\% of the 160 episodes, as reported in \cref{sec:exp:twostep}.

\method has a parameter count of 40\% of SmolVLA (0.25B vs.\ 0.6B) and is 3.4$\times$ times faster per inference call on the NVIDIA RTX 4050 laptop GPU (74\,ms vs.\ 254\,ms).
Because \method executes five actions per call, its cost per control step is just about 17\,ms.
LIBERO runs at 20\,Hz, so a call every five control steps has 250\,ms available, and \method runs in real time on this GPU.
On an RTX 4090, a considerably faster GPU than ours, TurboVLA~\cite{turbovla} measures one policy call (from input to action chunk, batch size 1) at 219.2\,ms for $\pi_0$, 212.7\,ms for SmolVLA, 112.2\,ms for OpenVLA-OFT, 87.3\,ms for VLA-Adapter and 31.2\,ms for TurboVLA; a \method call takes 74\,ms on the RTX 4050 laptop GPU.
Peak GPU memory of \method with a batch size of 1 is about 1.0\,GB compared to 1.2\,GB for SmolVLA.
As for training, the entire \method pipeline trains in about 8.5\,h on the same RTX 4050 laptop GPU.

\subsection{Two-step instructions}
\label{sec:exp:twostep}

\begin{table}[t]
\centering
\footnotesize
\setlength{\tabcolsep}{3.5pt}
\begin{tabular}{lrr}
\toprule
Method & Both steps & First step \\
\midrule
SmolVLA & 0/80 (0.0) & 0/80 \\
\textbf{\method} & \textbf{78/160 (48.8)} & 109/160 \\
\midrule
\multicolumn{3}{l}{\textit{\method, through the codes}} \\
\multicolumn{2}{l}{\quad stage~1 parses both steps correctly, in order} & 138/160 \\
\multicolumn{2}{l}{\quad both steps succeed when the parse is correct} & 74/138 \\
\multicolumn{2}{l}{\quad done code fires} & 156/160 \\
\multicolumn{2}{l}{\quad \ldots{} while step 1's goal holds} & 126/156 \\
\bottomrule
\end{tabular}
\caption{\textbf{Free-form two-step instructions} (16 task pairs; no two-step demonstrations, and no test pair in stage~1's training sentences). First step: its goal holds at the end of the episode. SmolVLA: 16 pairs $\times$ 5 episodes.}
\label{tab:twostep}
\end{table}

\begin{table}[t]
\centering
\footnotesize
\setlength{\tabcolsep}{4pt}
\begin{tabular}{lr}
\toprule
Two-step episode outcome (\method, 160) & Episodes \\
\midrule
Success & 78 \\
Parse error (stage 1) & 18 \\
First step not achieved & 11 \\
Done code fired before the first goal held & 5 \\
First step undone while executing the second & 22 \\
Second step not achieved, first step intact & 26 \\
\bottomrule
\end{tabular}
\caption{\textbf{Two-step failures by stage}, assigned in the order listed. No failure comes from a missed ``done'' code.}
\label{tab:twostepfail}
\end{table}

\method succeeds in two-step instructions, even though it was never trained on examples of two tasks performed together.
This differs from LIBERO-Long~\cite{libero}, which provides demonstrations of each multi-task.
Here, success therefore requires composing single-task skills from the instruction alone.
SmolVLA never completes the first goal (\cref{tab:twostep}) because the sentence is unlike any instruction that it was trained on.
\method completes both steps in 78 of 160 episodes.

Stage~1 splits 86.2\% of the sentences into the two canonical step codes in the right order (90\% of the fused sentences and 82.5\% of the joined \para paraphrases), including the sentences that name the second step first.
The model fires the done code for the first step in 156 out of 160 trials, and in 126 of those trials, the first goal was achieved.
Looking at the failures (\cref{tab:twostepfail}), switching between tasks is rarely the issue.
Misunderstanding the instructions accounts for 18 of the 82 failures.
Most failures occur during the second task, where the robot starts from scenes it has not seen during training; in 22 episodes, the robot undoes the first step while executing the second, and in 26, the second step fails with the first intact.
This is concentrated in the two pairs where the robot opens a drawer and then tries to move an object; the drawer is opened in all of the 20 episodes but closed again in 18, while the arm reaches for the bowl or the bottle (0/20 successes, against 8/10 when the second step is turning on the stove).
Refer to the supplementary for the per-pair results.

\subsection{Comparison with reported results}
\label{sec:exp:reported}

\begin{table}[t]
\centering
\footnotesize
\setlength{\tabcolsep}{1.8pt}
\begin{tabular}{lccccc}
\toprule
Method & Params & Goal & \para & Drop$\downarrow$ & PRIDE \\
\midrule
\multicolumn{6}{l}{\textit{Our protocol}} \\
\textbf{\method (ours)} & \textbf{0.25B} & 91.4 & \textbf{81.1} & \textbf{10.3} & \textbf{78.5} \\
\midrule
\multicolumn{6}{l}{\textit{Reported, original protocols}} \\
Xiaomi-Robotics-0~\cite{xiaomir0} & 4.7B & 98.8 & 76.0 & 22.8 & 69.2 \\
\pizerofive + GSR~\cite{paravla} & 3.6B & 91.0 & 75.6 & 15.4 & 70.4 \\
ParaVLA~\cite{paravla} & 0.33B & 92.0$^{a}$ & 72.5 & 19.5 & 66.9 \\
\pizerofive~\cite{pi05} & 3.3B & 97.6 & 71.4 & 26.2 & 65.4 \\
VLA-Adapter + GSR~\cite{paravla} & 0.5B & 98.0 & 70.9 & 27.1 & 62.0 \\
OpenVLA-OFT~\cite{openvlaoft} & 7.5B & 97.9 & 64.7 & 33.2 & 58.8 \\
X-VLA~\cite{xvla} & 0.9B & 97.8 & 62.1 & 35.7 & 52.7 \\
SmolVLA + GSR~\cite{paravla} & 0.45B & 78.0 & 49.1 & 28.9 & 41.4 \\
VLA-Adapter~\cite{vlaadapter} & 0.6B & 98.2 & 46.3 & 51.9 & 36.1 \\
SmolVLA~\cite{smolvla} & 0.45B & 72.0 & 4.5 & 67.5 & 2.6 \\
\bottomrule
\end{tabular}
\caption{\textbf{Full \para, compared with reported results} (success rate, \%).
Reported rows, including parameter counts, are copied from \cite{liberopara} and \cite{paravla} and use their own protocols; under ParaVLA's protocol SmolVLA scores 92 on Goal-100, against 83 under ours on the same initial states.
Drop = Goal $-$ \para. PRIDE~\cite{liberopara} weights each success by the paraphrase's distance from the original; ours is computed with the official script and per-paraphrase metadata. $^{a}$Goal-100.}
\label{tab:reported}
\end{table}

\Cref{tab:reported} compares \method with all results reported on the full \para benchmark.
\method achieves the highest success rate (81.1\%) and a PRIDE score of 78.5, with the smallest drop from the standard instructions (10.3 points), using only 0.25B parameters.
We evaluate each paraphrase only once, whereas the previous results were reported as the average over five seeds per paraphrase~\cite{liberopara}.
\Cref{sec:exp:main} provides the comparison under the same protocol.

First, models that achieve a success rate of 97--99\% on the standard \goal drop to 46--76\% on \para, showing that standard performance does not guarantee robustness to different wordings.
Second, the strongest results come from either larger models or approaches that decouple language understanding from action control (GSR, ParaVLA, and our \method).
TurboVLA~\cite{turbovla}, a strong model of similar size (0.2B), reportedly achieves 97.4\% on \goal but has not been evaluated on \para.

\subsection{What the codes reveal}
\label{sec:exp:attribution}

\begin{table}[t]
\centering
\footnotesize
\setlength{\tabcolsep}{4pt}
\begin{tabular}{lrr}
\toprule
\para episodes & Episodes & Success (\%) \\
\midrule
All & 4,092 & 81.1 \\
\quad step codes = canonical & 3,459 & 91.0 \\
\quad step codes $\neq$ canonical & 633 & 27.2 \\
\qquad codes of another \goal task & 181 & 0.0 \\
\qquad codes of no \goal task & 417 & 32.9 \\
\qquad canonical + spurious second step & 35 & 100.0 \\
\midrule
Action-expression paraphrases & 870 & 89.1 \\
Object-reference paraphrases & 259 & 84.9 \\
Compound paraphrases & 2,963 & 78.5 \\
\midrule
Standard instructions (\goal, 500) & 500 & 91.4 \\
\bottomrule
\end{tabular}
\caption{\textbf{Stage-1 codes and success on \para.} Stage~1 produces the canonical step codes for 84.5\% of paraphrases; when it does, success matches the standard-instruction rate.}
\label{tab:attribution}
\end{table}

Since both the interfaces use discrete codes, we can also trace failures to individual stages (\cref{tab:attribution}).
To identify which task a predicted plan executes, we first collapse consecutive repetitions and then pick the tasks from \goal whose demonstration chains are the closest in normalized edit distance over the first-level action codes.
This recovers the correct task for 93.7\% of the successful standard episodes.

\begin{figure}[t]
\centering
\includegraphics[width=\linewidth]{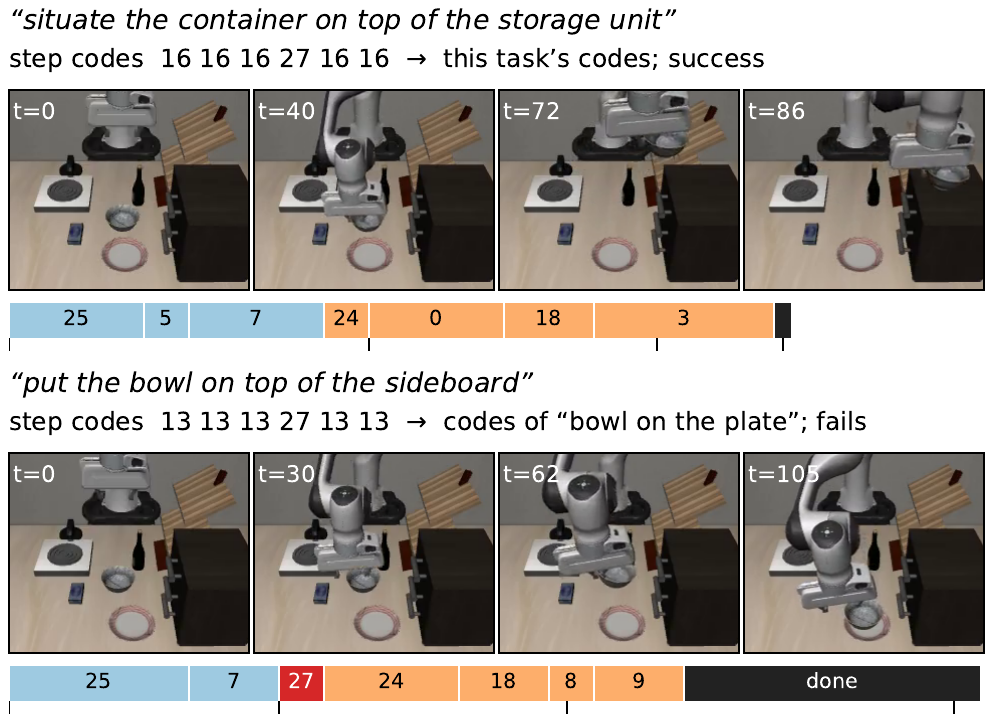}
\caption{\textbf{Two paraphrases of ``put the bowl on top of the cabinet''} (\para, initial states 29 and 49), with stage~1's step codes and the action codes the planner followed (blue: gripper open; red: closing; orange: closed; black: done code).
Top: the paraphrase receives the task's codes and the bowl is placed on the cabinet.
Bottom: \emph{sideboard} changes only the five goal slots (16 $\to$ 13), giving the codes of ``put the bowl on the plate''; the planner follows that task's chain and places the bowl on the plate.
Both episodes are re-runs of the deployed model and match the outcomes of the original \para run.}
\label{fig:rollouts}
\end{figure}

Stage~1 correctly produced the expected step codes for 3,459 of the 4,092 paraphrases (84.5\%).
On these episodes, \method achieved 91\%, which is almost identical to the 91.4\% on the standard instructions.
If we replace the outcome of the remaining 633 episodes with their corresponding standard-instruction outcomes (same task and initial state), the overall success rate was 91.3\%.
This suggests that fixing stage~1 alone could eliminate almost the entire performance gap between the standard and the paraphrase instructions.
The predicted codes also strongly determine the robot's behavior.
In 181 episodes, stage~1 generated the codes of a different \goal task, and none of these episodes succeeded.
In 175 episodes, the robot executed the incorrectly selected task.
For example, when the word ``cabinet'' was replaced with ``sideboard'' in ``put the bowl on top of the cabinet'', 32/34 paraphrases were interpreted as ``put the bowl on the plate'' (\cref{fig:rollouts}).
These errors are visible directly in the predicted codes before the robot takes any action.
Overall, understanding the errors accounts for 59.7\% of the \para failures, whereas failures on the standard instructions arise entirely from planning or execution.

\paragraph{Codes develop structure.}
Although stage~1 is trained only to agree across the paraphrases, an emergent meaningful structure is observed in the six codes.
Five slots represent the goal, while the fourth slot represents the object.
For example, the two tasks involving placing objects ``on top of the cabinet'' share goal code 16 but have different object codes (27 for the bowl and 17 for the wine bottle).

\paragraph{The action codes are readable.}
Although the action codebook is learned without any labels, its first-level codes correspond to recognizable robot motion primitives.
For example, Code 27 closes the gripper in 91\% of the segments, while codes 30 and 10 open it in all the segments.
Codes 25 and 7 move the open gripper forward and downward, while codes 0 and 18 lift and carry the objects with the gripper closed.
Pick-and-place demonstrations typically begin with the same approach and descent sequence (codes 25 and 7, sometimes through 5), but differ in how the object is carried and placed.
As a result, the planner's predicted code sequence can be read as a sequence of actions: approach, descent, grasp, lift, carry, and place.

\paragraph{Where to focus.}
Most of the failures on paraphrases are understanding errors, making stage~1 the clearest target for improvement.
Improving this stage requires only the text data.
Training the meaning codebook takes about an hour, and training the step coder takes about 5 minutes on an NVIDIA RTX 4050 laptop GPU.
Either component can be updated independently without retraining the planner or the actor or collecting any additional robot data.

\subsection{Design analyses}
\label{sec:exp:dev}

The development ablation in the supplementary material analyzes the contributions of different components in the final actor.
Temporal ensembling is the only individually statistically significant change ($p=0.04$).
Individually, fine-tuning the vision tail and the path-supervised attention each improves the success rate by one or two percentage points, although these gains are within statistical noise.
Combining them both gives the highest overall gain.

\paragraph{Discrete versus continuous step codes.}
We train a continuous planner using the same procedure and the data as the discrete planner.
Instead of receiving discrete step codes, the planner receives the meaning codebook's slot vectors before quantization.
On the \para subset, the discrete planner achieved 77\% compared to 65.5\% for the continuous planner ($p=0.002$).
On Goal-100, the discrete planner succeeded in 93 while the continuous planner succeeded in 82 ($p=0.03$).
These results suggest that the benefit of discrete codes goes beyond mapping different wordings of the same instruction to a common representation, consistent with the codes absorbing the imprecision of the small text model (\cref{sec:intro}).
However, each planner was trained with only one random seed.

\section{Limitations}
\label{sec:limitations}

Following ParaVLA~\cite{paravla}, we trained and evaluated \method only on \goal, as \para is constructed from this suite.
Furthermore, all the experiments were conducted in simulation.

One limitation of \para is that it contains paraphrases of only 10 underlying instructions.
A simpler approach could be to identify the closest matching instruction from the original set and pass it to an existing strong policy.
We haven't evaluated this alternative.

First, the multi-step execution relies on a scripted return to the starting pose between the steps.
Additionally, the second step can sometimes interfere with the outcomes of the first (\cref{sec:exp:twostep}).
Second, disabling the path-supervised attention at test time doesn't change the actor's validation loss, making the hypotheses hard to validate.
Finally, some initial states used during DAgger training overlap with those in the standard evaluation set.
We also report the results on the 300 evaluation episodes that exclude these overlapping states (\cref{sec:exp:main}).

\section{Conclusion}
\label{sec:conclusion}

We introduced \method, a framework that connects language understanding, planning, and robotic control through discrete, learnable codes.
Its architecture is chosen to run in real time on resource-constrained hardware: the instruction is encoded once by a small text encoder, and because the discrete codes absorb the errors of small models, every stage stays small, so that \method needs 74\,ms per inference call and 1.0\,GB of GPU memory on a laptop GPU.
Despite having only 250M params, \method outperforms SmolVLA on the standard \goal instructions under the same protocol and achieves the highest reported success rate on the \para benchmark.
The framework also supports free-form two-step instructions that SmolVLA cannot directly execute.
Our analyses show that most remaining failures on the paraphrased instructions originate from incorrect language understanding, which can be improved with the text data alone.
These findings highlight the value of decoupling language understanding from robotic control and suggest that VLA systems should be evaluated not only on their task success rate, but also by how effectively their failures can be understood, diagnosed, and addressed.

{
    \small
    \bibliographystyle{ieeenat_fullname}
    \bibliography{main}
}

\clearpage
\appendix
\renewcommand{\thesection}{\Alph{section}}
\setcounter{table}{0}\setcounter{figure}{0}
\renewcommand{\thetable}{S\arabic{table}}
\renewcommand{\thefigure}{S\arabic{figure}}
\makeatletter\newcommand{\gentab}[1]{\input{suppl/gen_#1.tex} }\makeatother
\setcounter{topnumber}{4}\setcounter{dbltopnumber}{4}\setcounter{totalnumber}{6}
\renewcommand{\topfraction}{0.95}\renewcommand{\dbltopfraction}{0.95}
\renewcommand{\textfraction}{0.05}
\renewcommand{\floatpagefraction}{0.85}\renewcommand{\dblfloatpagefraction}{0.85}
\makeatletter\setlength{\@fptop}{0pt}\setlength{\@dblfptop}{0pt}\makeatother
\newcommand{\paper}[1]{\cref{#1}}
\section*{Supplementary Material}
This supplement gives the implementation details and hyperparameters of every component (\cref{sec:s:impl}), the data used to train stage~1 and the two-step test sentences (\cref{sec:s:data}), per-task and per-paraphrase-type results (\cref{sec:s:pertask}), per-pair results on two-step instructions (\cref{sec:s:pairs}), and a closer look at both code vocabularies, including the comparison of the MDL action codebook with gripper-event segmentation (\cref{sec:s:codes}).
Section, table and equation numbers without an ``S'' refer to the main text.

\section{Implementation details}
\label{sec:s:impl}

\paragraph{Model size.}
The deployed model has 250.3M parameters: stage~1 (38.4M), the shared frozen SigLIP encoder (86.4M), the planner (26.1M) and the actor (99.4M).
The actor's count includes its own fine-tuned copy of the last four SigLIP layers (28.4M); since the planner still reads the frozen layers, both copies are deployed and both are counted.

\Cref{tab:s:hparams} lists the hyperparameters of every trained component; all are taken from the arguments stored with the deployed checkpoints.
All components use AdamW.
LIBERO task indices 10--19 are the ten \goal tasks; the planner and the actor are trained on their 428 demonstrations only, the action codebook on all 1,693 LIBERO demonstrations.

\begin{table*}[t]
\centering
\footnotesize
\setlength{\tabcolsep}{4pt}
\begin{tabular}{@{}lp{0.80\linewidth}@{}}
\toprule
Component & Hyperparameters \\
\midrule
Action codebook & 2 residual levels $\times$ 32 entries, code dim 64, MLP encoder/decoder (hidden 256); segment resampled to 16 steps; segment length 6--192; initial windows of 16 steps; 8 rounds $\times$ 3,000 autoencoder steps; EMA quantizers (decay 0.99) with dead-entry restarts \\
Meaning codebook & bge-base, top 2 layers fine-tuned (lr $2\times10^{-5}$); $M=6$ slots, shared cosine codebook $K_z=32$; slot transformer 2 layers, width 256, 4 heads, code dim 64; 6,000 steps, batch 256 pairs, lr $5\times10^{-4}$; decode-only warm-up 1,000 steps; loss weights swap 1, commitment 0.25, entropy 1, slot overlap 1 \\
Step coder (stage~1) & bge-small, top 4 layers fine-tuned (lr $5\times10^{-5}$), head lr $3\times10^{-4}$; up to 3 steps; 6,000 steps, batch 96; batch mix 35\% single-step, 20\% fused two-step, 45\% joined two-step; half of joined sentences pair two \goal tasks; presence loss weight 1 \\
Planner & 6 layers, width 512, 8 heads, 26.1M parameters; memory 2$\times$64 SigLIP tokens + state + 6 step codes; chain of 3 segments $\times$ 2 levels + end token; 3,000 steps, batch 32, lr $3\times10^{-4}$, weight decay 0.01, warm-up 300; DAgger fraction 0.5; step-code substitution 0.05; image shift 8 px \\
Planner, done fine-tuning & 1,000 steps, lr $10^{-4}$, warm-up 100; 16 two-step states per batch, done loss weight 1 \\
Actor & 12 layers, width 512, 8 heads; adaLN-zero conditioning on the flow time and 6 code slots; chunk 50, 5 executed; 2$\times$64 pooled SigLIP tokens + glimpses over 2$\times$256 patches (2 context blocks); last 4 SigLIP layers fine-tuned at 0.2$\times$ lr; batch 32, weight decay $10^{-4}$, EMA 0.999; null codes 0.1, level-2 dropout 0.3; planner codes 0.5 (from stage 3); path loss weight 0.1, $\sigma=12$ px; 5 stages, 39k steps (lr $2\times10^{-4}$ then $5\times10^{-5}$) \\
Inference & planner and actor every 5 control steps; greedy chain decoding; 10 Euler flow steps; temporal ensembling: decay 0.9, horizon 20 steps, gripper from the newest chunk; switch after 3 consecutive done calls \\
\bottomrule
\end{tabular}
\caption{\textbf{Hyperparameters} of the deployed model, from the arguments stored with each checkpoint.}
\label{tab:s:hparams}
\end{table*}

\paragraph{Action codes.}
A segment's 7-D actions (6 end-effector deltas and the gripper command, normalized per dimension) are resampled to 16 steps and concatenated with the segment's log-duration; an MLP encoder maps them to a 64-D vector that two residual EMA vector quantizers (32 entries each, dead entries restarted) encode, and an MLP decoder reconstructs the 16 steps, which are resampled back to the segment's true length so that the cost in \cref{eq:mdl} is additive in original time.
The reconstruction term uses a Gaussian whose per-dimension variance is re-estimated every round; the code cost is $\log 32$ nats for the second level and $\log(192-6+1)$ for the duration (segments are 6 to 192 steps long), and the first-level bigram prior is estimated per task with Laplace smoothing on the training demonstrations.
Each of the 8 rounds re-segments every demonstration by dynamic programming and then trains the autoencoder for 3,000 steps on the new segments.
10\% of the demonstrations of every task are held out from autoencoder training to measure reconstruction (\cref{sec:s:codebook}); all 1,693 are segmented.

\paragraph{Meaning codebook.}
It is trained on paraphrase pairs of 231 instructions (40 LIBERO and 191 recombined, \cref{sec:s:data}); 28 of the recombined instructions are held out entirely to test generalization to unseen meanings, and 10\% of the paraphrases of the others are held out for validation, leaving 21,880 training sentences.
Each step samples two different sentences of one instruction, $\ell_A$ and $\ell_B$.
A two-layer decoder generates $\ell_B$'s tokens from $\ell_A$'s six quantized codes; a swap loss makes $\ell_A$'s slot logits predict $\ell_B$'s codes and vice versa, enabled after 1,000 decode-only steps so that the codes first carry meaning; a commitment loss, a per-slot batch-entropy bonus against codebook collapse and a penalty on the overlap of the code distributions of different slots complete the objective.
The lower layers of bge-base are frozen and their outputs cached; the top two are fine-tuned.
After training, the 10 \goal instructions receive 10 distinct code tuples (\cref{tab:s:canon}).

\paragraph{Step coder.}
The step coder replaces bge-base with bge-small (top four layers fine-tuned), keeps the meaning codebook frozen, and gives the six slot queries a learned per-step embedding initialized at zero, so that at initialization every step reproduces the single-step codes.
It predicts up to three steps and a presence probability per step; the number of steps is the index of the first absent step.
A training batch mixes single-step paraphrases (35\%), LLM-fused two-step rewrites of training pairs (20\%) and two-step sentences joined on the fly from paraphrases of two steps (45\%); half of the joined sentences pair two \goal tasks.
The targets are the meaning-codebook codes of each step's canonical instruction, with a cross-entropy loss per slot and a binary cross-entropy loss on presence.

\begin{table*}[t]
\centering
\footnotesize
\setlength{\tabcolsep}{4.5pt}
\begin{tabular}{rlcccccccccc}
\toprule
& & \multicolumn{2}{c}{\goal, 50 per task} & \multicolumn{7}{c}{Goal-100, initial states 0--9 (10 per task)} \\
\cmidrule(lr){3-4}\cmidrule(lr){5-11}
& Task & \method & SmolVLA & seed 0 & seed 1 & seed 2 & SmolVLA & +para LoRA & discrete & continuous \\
\midrule
0 & open middle drawer & 48 & 40 & 10 & 10 & 10 & 9 & 9 & 10 & 10 \\
1 & open top drawer, bowl inside & 46 & 26 & 10 & 7 & 10 & 6 & 4 & 10 & 9 \\
2 & push plate to stove front & 40 & 45 & 9 & 10 & 9 & 9 & 8 & 8 & 9 \\
3 & bowl on plate & 48 & 46 & 10 & 10 & 10 & 10 & 6 & 10 & 9 \\
4 & bowl on stove & 50 & 45 & 10 & 10 & 10 & 10 & 6 & 10 & 9 \\
5 & bowl on top of cabinet & 48 & 47 & 10 & 10 & 9 & 9 & 6 & 10 & 9 \\
6 & cream cheese in bowl & 43 & 33 & 7 & 6 & 6 & 5 & 4 & 7 & 7 \\
7 & wine bottle on rack & 39 & 42 & 9 & 9 & 8 & 7 & 6 & 9 & 7 \\
8 & wine bottle on top of cabinet & 45 & 40 & 9 & 8 & 8 & 8 & 9 & 9 & 3 \\
9 & turn on stove & 50 & 50 & 10 & 10 & 10 & 10 & 9 & 10 & 10 \\
\midrule
 & \textit{All} & 457/500 & 414/500 & 94 & 90 & 90 & 83 & 67 & 93 & 82 \\
 
\bottomrule
\end{tabular}
\caption{\textbf{\goal per task} (successes). Seed 0 is the deployed model. Discrete and continuous are the two planners of the paper's discrete-versus-continuous comparison (same actor). \emph{Generated from the run logs.}}
\label{tab:s:goal}
\end{table*}

\begin{table*}[t]
\centering
\footnotesize
\setlength{\tabcolsep}{4pt}
\begin{tabular}{rrccccccccccc}
\toprule
& \multicolumn{7}{c}{\para full, \method} & \multicolumn{5}{c}{\para subset (200), successes} \\
\cmidrule(lr){2-8}\cmidrule(lr){9-13}
Task & Episodes & Success & Canon. & Succ.\,$\mid$\,canon. & Und. & Plan. & Exec. & \method & SmolVLA & +para LoRA & discrete & continuous \\
\midrule
0 & 416 & 94.2 & 96.9 & 94.3 & 1 & 7 & 16 & 19/20 & 3/20 & 5/20 & 19/20 & 19/20 \\
1 & 410 & 64.4 & 72.2 & 87.8 & 110 & 4 & 32 & 7/20 & 1/20 & 3/20 & 13/20 & 11/20 \\
2 & 406 & 87.9 & 91.6 & 88.4 & 6 & 13 & 30 & 16/20 & 0/20 & 12/20 & 16/20 & 14/20 \\
3 & 386 & 87.8 & 78.2 & 99.3 & 45 & 2 & 0 & 16/19 & 1/19 & 5/19 & 17/19 & 16/19 \\
4 & 403 & 94.0 & 85.9 & 98.0 & 17 & 0 & 7 & 18/20 & 1/20 & 4/20 & 16/20 & 14/20 \\
5 & 410 & 81.5 & 80.0 & 96.3 & 64 & 2 & 10 & 18/20 & 3/20 & 6/20 & 14/20 & 14/20 \\
6 & 411 & 67.4 & 85.9 & 75.1 & 46 & 47 & 41 & 14/20 & 0/20 & 2/20 & 11/20 & 12/20 \\
7 & 413 & 70.0 & 86.9 & 80.5 & 54 & 8 & 62 & 13/20 & 0/20 & 7/20 & 14/20 & 7/20 \\
8 & 423 & 66.4 & 71.6 & 90.1 & 112 & 7 & 23 & 16/21 & 3/21 & 11/21 & 14/21 & 6/21 \\
9 & 414 & 98.6 & 95.9 & 100.0 & 6 & 0 & 0 & 20/20 & 4/20 & 18/20 & 20/20 & 18/20 \\
\midrule
\textit{All} & 4,092 & 81.1 & 84.5 & 91.0 & 461 & 90 & 221 & 157/200 & 16/200 & 73/200 & 154/200 & 131/200 \\
 
\bottomrule
\end{tabular}
\caption{\textbf{\para per task.} Success, Canon.\ (stage~1 produced the task's canonical step codes) and Succ.\,$\mid$\,canon.\ (success when it did) are percentages; Und./Plan./Exec.\ count failures as understanding (stage-1 codes differ from the canonical ones), planning (codes correct, but the action-code chain the planner followed is closest to the demonstrations of another task) or execution (codes and chain match the task, but the episode fails), as in \paper{sec:exp:attribution}. Task indices as in \cref{tab:s:goal}. \emph{Generated from the run logs.}}
\label{tab:s:para}
\end{table*}

\paragraph{Planner.}
The planner's memory holds the SigLIP tokens of both cameras ($256\times256$ input, average-pooled to $8\times8$ tokens per camera), the robot state and the six step codes; a 6-layer decoder generates the action-code chain greedily, two tokens (levels 1 and 2) per segment for three segments, followed by an end-of-chain token.
In training, each frame's step codes are those of a random paraphrase of its task, and each code is replaced by a random one with probability 0.05; images are randomly shifted by up to 8 pixels.
The behavior-cloned planner is trained for 3,000 steps with half of every batch drawn from the DAgger states.
They come from 200 rollouts of an earlier behavior-cloned planner (20 per task): from initial states 10--19 the planner's codes are always executed, and from initial states 20--29 each call executes the labeling oracle's codes with probability 0.5.
Every state at which the planner was called is recorded and labeled with the code of the nearest training frame of the same task, giving 10,601 states.
The done fine-tuning then trains for 1,000 more steps, adding to every batch 16 states from the 80 two-step rollouts of \paper{sec:method:planner} (9,795 states, 1,299 of them labeled done).

\paragraph{Actor.}
The actor is trained in five consecutive stages, each initialized from the previous one (39k steps in total): 15k steps with ground-truth codes at learning rate $2\times10^{-4}$, then at $5\times10^{-5}$ 4k steps with ground-truth codes, 5k steps in which each sample is conditioned on the planner's predicted codes with probability 0.5, 5k steps that add path-supervised code attention, and 10k steps that add the fine-tuned SigLIP tail.
Throughout, all six code slots are replaced by a null code with probability 0.1 and each second-level code is dropped with probability 0.3.
The code attention is a set of glimpses: after two context blocks over the image tokens, the state and the codes, each code slot forms one query per camera over the full $16\times16$ patch grid, and what it reads is added to that slot's token.
The auxiliary loss is a cross-entropy between the front camera's attention map of a slot and a target map built from the gripper's projected image path during the segment that the slot covers: a sum of Gaussians ($\sigma=12$ pixels) over the path, with half of the mass on the segment's end point, weighted by 0.1.
The attention follows the path (on held-out frames, 80\% of a slot's attention falls on the 10 patches that the target ranks highest, against 4\% for uniform attention), but the actor barely uses what it reads: setting the glimpses to zero changes the held-out flow loss of the deployed actor from 0.493 to 0.488.
This agrees with the development ablation (\cref{tab:s:dev}), where adding the path-supervised attention changes success by one episode out of 200.
Flow matching follows SmolVLA: $x_\tau = \tau\epsilon + (1-\tau)a$ with $\tau\sim\mathrm{Beta}(1.5,1)$, the network predicts $\epsilon-a$, and inference integrates 10 Euler steps from $\tau=1$.

\paragraph{Inference and two-step execution.}
Every 5 control steps the planner is called once and the actor once (with its 10 flow steps) on a 50-step chunk, of which 5 steps are executed with temporal ensembling (\paper{sec:method:actor}).
For two-step instructions, the switch happens when the planner's chain has started with the end-of-chain token in three consecutive calls.
The scripted return then runs as ordinary environment steps within the 980-step budget: the gripper opens for 15 steps, the hand lifts 8\,cm, and an operational-space controller brings it back to its episode-start position and orientation (to within 1\,cm and 0.05\,rad, at most 150 steps per move).

\paragraph{SmolVLA + paraphrase LoRA.}
LoRA adapters of rank 32 on the language model and the action expert of the public SmolVLA LIBERO checkpoint (learning rate $10^{-4}$) are trained for 3,000 steps with batch size 8 and 4 accumulation steps on the 428 \goal demonstrations, with the instruction replaced by a random paraphrase from the training paraphrases (\cref{sec:s:data}) with probability 0.8.
It runs with SmolVLA's released inference settings (one action executed per call).

\paragraph{Compute and measurements.}
All training and evaluation ran on one laptop with an NVIDIA RTX 4050 Laptop GPU (6\,GB), an Intel Core i7-13620H and 15\,GB of RAM, with PyTorch 2.11, LeRobot 0.6.1, MuJoCo 3.8.1 and robosuite 1.4.0.
\Cref{tab:s:compute} breaks down the 8.5 GPU-hours of training.
Latency is the CUDA-synchronized time of each policy call at batch size 1 on an otherwise idle GPU, logged over one episode of every \goal task; peak GPU memory is the maximum allocated from the third policy call on, after lazy initialization.
The released SmolVLA checkpoint also contains its language model and action expert, which \method does not use; they are moved off the GPU for the memory measurement (1.88\,GB with them resident).

\begin{table}[t]
\centering
\footnotesize
\begin{tabular}{lr}
\toprule
Component & GPU-hours \\
\midrule
Actor (5 stages, 39k steps) & 5.45 \\
Planner (behavior cloning + DAgger, done fine-tuning) & 0.50 \\
DAgger rollouts (200 episodes) & 0.59 \\
Two-step rollouts for the done labels (80 episodes) & 0.61 \\
Meaning codebook & $\approx$1.0 \\
Action codebook (MDL, 8 rounds) & $\approx$0.2 \\
Step coder & 0.08 \\
\midrule
Total & $\approx$8.5 \\
\bottomrule
\end{tabular}
\caption{\textbf{Training compute} on the 6\,GB laptop GPU, from the training logs. Excluded: development runs and evaluation.}
\label{tab:s:compute}
\end{table}

\paragraph{Development ablation.}
\Cref{tab:s:dev} traces the final actor on the selection sets (\paper{sec:exp:dev}).

\begin{table}[t]
\centering
\footnotesize
\setlength{\tabcolsep}{4pt}
\begin{tabular}{lcc}
\toprule
Configuration (selection sets, 200 eps.) & Success & $p$ vs.\ prev. \\
\midrule
Actor, executed without ensembling & 161 (80.5) & -- \\
+ temporal ensembling & 176 (88.0) & 0.04 \\
+ fine-tuned SigLIP tail & 180 (90.0) & 0.57 \\
+ path-supervised code attention (final) & 182 (91.0) & 0.82 \\
\midrule
Control: + training steps only (with ens.) & 173 (86.5) & -- \\
\bottomrule
\end{tabular}
\caption{\textbf{Development ablation} on \goal initial states 0--9 and 30--39 (successes out of 200, \%), all with the same stage~1 and planner.
These are the selection sets, so the confirmation on untouched initial states 40--49 (89\%) is the unbiased estimate.}
\label{tab:s:dev}
\end{table}

\section{Data}
\label{sec:s:data}

\paragraph{Paraphrases.}
Training paraphrases were generated with Claude Sonnet through its command-line interface, from the instruction alone (no images).
First, the model wrote new instructions that recombine the actions, objects and places of the 40 LIBERO training instructions; after removing exact repeats this gave 191 recombined instructions, for which no demonstrations exist and which only serve to teach the meaning codebook to factor meaning.
Then, for each of the 231 instructions, two calls with different emphases asked for 40 paraphrases each, and one more round asked for paraphrases that refer to every object and place by a description.
The paraphrase prompt was:
\begin{quote}\footnotesize\itshape
You write training data for a robot that follows spoken instructions in a tabletop kitchen scene. Rewrite the instruction below in exactly $n$ different ways that a person might say it. Every rewrite must ask for exactly the same action on exactly the same objects and locations; do not add, drop or change any object, location, count or step. [emphasis] Keep each one a single sentence or two short ones. Output ONLY a JSON array of $n$ strings, nothing else.
\end{quote}
The two emphases asked for (i) varied everyday wording (synonyms, reordered clauses, polite requests, questions, indirect requests, casual speech) and (ii) less common wording (descriptions of objects and places by colour, shape, material, use or position, uncommon verbs, extra harmless context, hints and statements of need).
After deduplication there are 27,222 paraphrases, 4,652 of them for the 40 LIBERO instructions and 105--117 for each \goal instruction.
The 74 that exactly matched a \para paraphrase (after lowercasing and removing punctuation) were removed; \para was never shown to the generator.
Sample paraphrases of ``put the bowl on top of the cabinet'': \emph{``Bowl goes on top of the cabinet.''}, \emph{``Set the curved food container on the upper surface of the cupboard.''}, \emph{``Place the round dish you'd eat soup or cereal from onto the upper surface of the storage unit.''}

\begin{table*}[t]
\centering
\scriptsize
\setlength{\tabcolsep}{2.2pt}
\begin{tabular}{rlcccc@{\hspace{8pt}}rlcccc}
\toprule
& Step 1 $\to$ step 2 & Parse & First & Both & SmolVLA & & Step 1 $\to$ step 2 & Parse & First & Both & SmolVLA \\
\midrule
0 & open middle drawer $\to$ bowl on stove & 10 & 0 & 0 & 0/5 & 8 & cream cheese in bowl $\to$ bowl on plate & 7 & 6 & 2 & 0/5 \\
1 & open middle drawer $\to$ wine bottle on rack & 10 & 2 & 0 & 0/5 & 9 & cream cheese in bowl $\to$ turn on stove & 7 & 4 & 4 & 0/5 \\
2 & open middle drawer $\to$ turn on stove & 10 & 10 & 8 & 0/5 & 10 & wine bottle on rack $\to$ bowl on plate & 7 & 5 & 4 & 0/5 \\
3 & push plate to stove front $\to$ wine bottle on rack & 10 & 8 & 4 & 0/5 & 11 & wine bottle on rack $\to$ bowl on stove & 7 & 5 & 4 & 0/5 \\
4 & bowl on plate $\to$ wine bottle on rack & 8 & 7 & 0 & 0/5 & 12 & wine bottle on rack $\to$ cream cheese in bowl & 9 & 6 & 3 & 0/5 \\
5 & bowl on plate $\to$ turn on stove & 10 & 10 & 8 & 0/5 & 13 & wine bottle on top of cabinet $\to$ bowl on stove & 8 & 8 & 8 & 0/5 \\
6 & bowl on stove $\to$ turn on stove & 8 & 10 & 10 & 0/5 & 14 & turn on stove $\to$ bowl on stove & 8 & 9 & 6 & 0/5 \\
7 & bowl on top of cabinet $\to$ turn on stove & 9 & 9 & 9 & 0/5 & 15 & turn on stove $\to$ wine bottle on rack & 10 & 10 & 8 & 0/5 \\
\midrule
\multicolumn{8}{r}{\textit{All 16 pairs}} & 138/160 & 109/160 & 78/160 & 0/80 \\
 
\bottomrule
\end{tabular}
\caption{\textbf{Two-step results per pair} (\method: 10 episodes per pair, initial states 0--9; SmolVLA: 5). Parse: stage~1 produced both canonical step codes in order. First: step~1's goal holds at the end of the episode. Both: success. \emph{Generated.}}
\label{tab:s:pairs}
\end{table*}

\begin{table*}[t]
\centering
\footnotesize
\begin{tabular}{lrrl}
\toprule
Instruction's task $\to$ task whose codes stage~1 produced & Episodes & Planner follows it & Example paraphrase \\
\midrule
bowl on top of cabinet $\to$ bowl on plate & 48 & 46 & \emph{gently put the vessel on top of the sideboard} \\
wine bottle on top of cabinet $\to$ wine bottle on rack & 41 & 41 & \emph{swiftly put the wine container on top of the sideboard} \\
wine bottle on rack $\to$ wine bottle on top of cabinet & 38 & 38 & \emph{now put the wine bottle on the kitchen rack} \\
wine bottle on top of cabinet $\to$ bowl on top of cabinet & 24 & 21 & \emph{deliberately put the vessel on top of the cupboard} \\
open top drawer, bowl inside $\to$ bowl on top of cabinet & 7 & 6 & \emph{the top bin is the spot for the vessel} \\
bowl on stove $\to$ turn on stove & 6 & 6 & \emph{the stove is still clear} \\
turn on stove $\to$ bowl on stove & 6 & 6 & \emph{the food is ready to be cooked on the stove} \\
\midrule others & 11 & 11 & \\
\textit{All} & 181 & 175 & \\
 
\bottomrule
\end{tabular}
\caption{\textbf{Most frequent stage-1 confusions on \para}: paraphrases that received exactly the canonical step codes of another \goal task. None of these 181 episodes succeeds. Planner follows it: the action-code chain the planner followed is closest to the demonstrations of that other task (\paper{sec:exp:attribution}). \emph{Generated.}}
\label{tab:s:confusions}
\end{table*}

\paragraph{Two-step sentences.}
The 16 test pairs are ordered pairs of \goal tasks whose two goals can hold at the same time in the shared scene (\cref{tab:s:pairs}).
Every unordered pair of steps that forms a test pair is excluded from all stage-1 training data, in either order.
Training uses two kinds of two-step sentences built from the other pairs.
\emph{Joined} sentences connect two paraphrases with one of 19 templates, three of which name the second step first (\eg ``before you $B$, $A$'', ``$B$ after you $A$''); 8,986 are precomputed, and more are formed on the fly.
\emph{Fused} sentences (1,528) are single-sentence rewrites written by Claude Haiku, which was asked for both steps in one or two sentences with the order unambiguous, varied wording, pronouns where natural, and occasionally the second step mentioned first.
The test sentences are built the same way from the 16 held-out pairs: for even initial states a fused sentence (482 generated, \eg ``Pull the cabinet's middle drawer out and then move the wine bottle onto the rack''), and for odd initial states two \para paraphrases of the two steps joined by a template (640 generated, \eg ``Position the serving bowl on the stove after you open the middle compartment of the cupboard'').

\begin{table*}[t]
\centering
\footnotesize
\scriptsize
\setlength{\tabcolsep}{2.2pt}
\begin{tabular}{rrrlrrrlrl@{\hspace{10pt}}rrrlrrrlrl}
\toprule
Code & Segs. & Len. & Gripper & $\Delta x$ & $\Delta y$ & $\Delta z$ & Motion & Tasks & Main & Code & Segs. & Len. & Gripper & $\Delta x$ & $\Delta y$ & $\Delta z$ & Motion & Tasks & Main \\
\midrule
25 & 279 & 16 & open & +10.6 & +2.0 & -1.1 & fwd & 7 & 3,4,5 & 29 & 100 & 14 & closed & +4.3 & -3.1 & -2.9 & fwd, $-y$ & 7 & 8 \\
23 & 268 & 16 & open & +2.6 & -0.3 & -1.7 & small & 3 & 0,7 & 19 & 89 & 12 & open & -2.3 & -1.6 & -5.0 & down & 8 & 1,0 \\
20 & 261 & 9 & open & +0.2 & +0.4 & +1.5 & small & 10 & 7 & 30 & 80 & 10 & opens & +0.3 & +0.2 & +1.1 & small & 8 & 5,3,6 \\
7 & 247 & 15 & open & -2.6 & -0.4 & -12.4 & down & 9 & 4,3,5 & 2 & 65 & 10 & closed & -0.6 & -1.2 & -4.8 & down & 7 & 7,0 \\
18 & 233 & 14 & closed & +7.7 & -4.3 & +10.6 & fwd, $-y$, up & 6 & 8,3,5,1 & 15 & 65 & 11 & closed & +0.2 & +0.9 & +2.8 & small & 9 & 8,4 \\
6 & 219 & 11 & open & -1.5 & -0.5 & -5.5 & down & 10 & 1,7 & 16 & 63 & 18 & closed & +1.3 & -7.7 & +7.7 & $-y$, up & 3 & 7 \\
24 & 206 & 10 & closed & -0.4 & -0.1 & -0.3 & small & 8 & 8,4,3 & 14 & 56 & 13 & open & -1.6 & -3.4 & -3.1 & $-y$, down & 4 & 7,0 \\
12 & 195 & 24 & open & -4.0 & +18.0 & -10.2 & back, $+y$, down & 5 & 2,9,1,0 & 22 & 53 & 12 & closed & -1.8 & -0.7 & -7.9 & down & 5 & 3,1 \\
0 & 168 & 15 & closed & +0.4 & +0.0 & +7.4 & up & 8 & 5,1,6,7 & 8 & 51 & 11 & closed & +7.7 & -2.5 & +0.0 & fwd & 4 & 3,1 \\
9 & 148 & 10 & closed & -2.1 & +0.2 & -4.9 & down & 8 & 4,3 & 21 & 48 & 11 & open & +0.1 & -5.0 & -1.9 & $-y$ & 5 & 0,7 \\
31 & 147 & 20 & open & -11.8 & +1.9 & -7.8 & back, down & 6 & 9,2,1 & 26 & 45 & 12 & closed & -2.1 & +4.9 & +7.7 & $+y$, up & 3 & 4 \\
5 & 137 & 11 & open & +5.6 & -0.7 & -6.0 & fwd, down & 6 & 1,2,5 & 10 & 39 & 10 & opens & -0.8 & -0.3 & -1.7 & small & 7 & 4,6,7,3 \\
28 & 121 & 11 & open & +1.9 & -0.3 & +2.1 & small & 10 & 8 & 11 & 32 & 14 & closed & -2.9 & +6.2 & -5.3 & $+y$, down & 2 & 4 \\
17 & 119 & 14 & closed & -0.8 & +0.4 & -4.7 & down & 2 & 9 & 4 & 31 & 11 & open & +1.3 & -0.1 & -4.1 & down & 6 & 1,7,0 \\
27 & 117 & 12 & closes & -3.0 & +1.7 & -6.0 & down & 9 & 3,9 & 13 & 30 & 12 & closed & -5.1 & +8.9 & +1.2 & back, $+y$ & 1 & 4 \\
3 & 106 & 16 & closed & +0.7 & -8.5 & -0.6 & $-y$ & 5 & 5,6 & 1 & 23 & 9 & closed & -0.6 & -0.8 & -2.2 & small & 7 & 7,1 \\
 
\bottomrule
\end{tabular}
\caption{\textbf{Dictionary of the first-level action codes}, over the segments they label in the \goal demonstrations, sorted by frequency.
Segs.: number of segments; Len.: mean segment length in control steps.
Gripper: closed or open throughout most segments, or closes/opens within at least half of them.
$\Delta x, \Delta y, \Delta z$: mean summed end-effector command over the segment (normalized action units; $x$ forward, $z$ up); Motion names the components with magnitude $\geq 3$.
Tasks: number of \goal tasks that use the code; Main: those with at least 15\% of its segments (indices as in \cref{tab:s:goal}). \emph{Generated.}}
\label{tab:s:dict}
\end{table*}

\begin{table*}[t]
\centering
\footnotesize
\setlength{\tabcolsep}{4pt}
\begin{tabular}{lcccccccc}
\toprule
& Median & Median & Motion & \multicolumn{2}{c}{Reuse across tasks (\%)} & Within-task & Task ID & Gripper changes \\
\cmidrule(lr){5-6}
Segmentation & segment & chain length & rec.\ error$\downarrow$ & 2-grams & 3-grams & NED$\downarrow$ & from chain (\%) & with a boundary (\%) \\
\midrule
gripper events (hand-designed) & -- & 3 & 0.355 & 72 & 35 & 0.33 & 73 & 100$^{*}$ \\
MDL, 16-step init (\textbf{ours}) & 13 & 9 & 0.188 & 96 & 74 & 0.46 & 62 & 87 \\
MDL, 32-step init & 16 & 7 & 0.231 & 95 & 71 & 0.56 & 61 & 72 \\
MDL, 64-step init & 38 & 3 & 0.328 & 88 & 61 & 0.50 & 59 & 26 \\
 
\bottomrule
\end{tabular}
\caption{\textbf{Action codebooks compared} (2 levels $\times$ 32 entries each), on held-out demonstrations.
Chain length counts first-level codes after merging repeats.
Motion rec.\ error: per-frame squared error of the 6 normalized motion dimensions, decoding every segment from its codes (the gripper is excluded because it is constant within gripper-event segments).
Reuse: fraction of first-level 2- and 3-grams that also occur in another task's chains.
Within-task NED: normalized edit distance between chains of the same task.
Task ID: a held-out demonstration's task is that of the nearest training chain.
Gripper changes: within 3 steps of a segment boundary; $^{*}$by construction. \emph{Generated.}}
\label{tab:s:codebook}
\end{table*}

\section{Per-task results}
\label{sec:s:pertask}

\Cref{tab:s:goal,tab:s:para} give the per-task results behind \paper{sec:exp:main}, \paper{sec:exp:attribution} and the discrete-versus-continuous comparison; \cref{tab:s:paratype} splits the full \para results by the benchmark's paraphrase types.

Three tasks account for most understanding errors: ``open the top drawer and put the bowl inside'' (110), and the two ``on top of the cabinet'' tasks (64 and 112), whose paraphrases often call the cabinet a \emph{sideboard}, \emph{hutch} or \emph{cupboard} (\cref{tab:s:confusions}).
When stage~1 gets the codes right, success is at least 87.8\% on every task except two of the three weakest tasks on standard instructions (\cref{tab:s:goal}), ``put the cream cheese in the bowl'' (75.1\%) and ``put the wine bottle on the rack'' (80.5\%).

\begin{table}[t]
\centering
\footnotesize
\setlength{\tabcolsep}{3pt}
\begin{tabular}{lrrr}
\toprule
\para type & Episodes & Success & Canonical \\
\midrule
\multicolumn{4}{l}{\textit{Action expression varied (object named as in \goal)}} \\
\quad lexical: addition/deletion & 100 & 93.0 & 100.0 \\
\quad lexical: contextual synonym & 79 & 91.1 & 100.0 \\
\quad lexical: habitual synonym & 74 & 85.1 & 100.0 \\
\quad pragmatic: embedded imperative & 93 & 95.7 & 100.0 \\
\quad pragmatic: hint & 88 & 81.8 & 92.0 \\
\quad pragmatic: need statement & 93 & 92.5 & 100.0 \\
\quad pragmatic: permission directive & 83 & 91.6 & 100.0 \\
\quad pragmatic: question directive & 87 & 77.0 & 90.8 \\
\quad structural: coordination & 98 & 89.8 & 87.8 \\
\quad structural: subordination & 75 & 92.0 & 100.0 \\
\midrule \multicolumn{4}{l}{\textit{Object reference varied (action as in \goal)}} \\
\quad lexical: addition/deletion & 98 & 90.8 & 96.9 \\
\quad lexical: contextual synonym & 87 & 78.2 & 85.1 \\
\quad lexical: habitual synonym & 74 & 85.1 & 89.2 \\
\midrule \multicolumn{4}{l}{\textit{Compound: object variation $\times$ action variation family}} \\
\quad addition/deletion $\times$ lexical & 300 & 90.0 & 95.0 \\
\quad addition/deletion $\times$ pragmatic & 497 & 84.1 & 90.7 \\
\quad addition/deletion $\times$ structural & 200 & 85.0 & 85.5 \\
\quad contextual synonym $\times$ lexical & 300 & 73.3 & 74.3 \\
\quad contextual synonym $\times$ pragmatic & 490 & 70.8 & 69.8 \\
\quad contextual synonym $\times$ structural & 199 & 68.8 & 69.3 \\
\quad habitual synonym $\times$ lexical & 298 & 80.9 & 83.6 \\
\quad habitual synonym $\times$ pragmatic & 488 & 75.8 & 75.8 \\
\quad habitual synonym $\times$ structural & 191 & 79.6 & 79.6 \\
 
\bottomrule
\end{tabular}
\caption{\textbf{Full \para by paraphrase type} (\method, \%). Canonical: stage~1 produced the task's canonical step codes. Objects vary only lexically in \para. \emph{Generated.}}
\label{tab:s:paratype}
\end{table}

Paraphrases that only change how the action is expressed are almost always understood (87.8--100\% canonical codes); the questions and hints are the hardest of them (77.0\% and 81.8\% success).
Changing the name of an object by a contextual synonym (\eg \emph{vessel} for bowl, \emph{sideboard} for cabinet) is the most damaging change, alone (78.2\%) or combined with any change of the action expression (68.8--73.3\%), and in the compound rows success is within 7 points of the canonical-code rate, which again places the remaining gap in stage~1.

\section{Two-step instructions per pair}
\label{sec:s:pairs}

\Cref{tab:s:pairs} gives the results behind \paper{tab:twostep}.
Pairs whose second step is turning on the stove are the easiest (8--10 of 10 for four of the five).
The two pairs that open the drawer and then move an object fail in all 20 episodes: the drawer is opened in every episode but closed again in 18, as the arm sweeps past it towards the bowl or the bottle (\paper{sec:exp:twostep}).
``Put the bowl on the plate, then the wine bottle on the rack'' also fails in all 10 episodes, although step~1 is completed in 7: the second step starts from a scene with the bowl on the plate, which the rack task's demonstrations never start from.
Parse errors concentrate in the pairs that start with the cream cheese or the wine bottle on the rack (3 of 10 wrong in four of these five pairs); SmolVLA achieves neither step in any of the 80 episodes.

\section{The codes}
\label{sec:s:codes}

\paragraph{Step codes.}
\Cref{tab:s:canon} lists the step codes of the 10 canonical instructions.
Five slots repeat one goal code and the fourth slot holds an object code (\paper{sec:exp:attribution}).
\Cref{tab:s:confusions} lists the most frequent cases in which a paraphrase received exactly the step codes of another \goal task; in 175 of these 181 episodes the planner then followed the chain of that other task.

\begin{table}[t]
\centering
\footnotesize
\setlength{\tabcolsep}{4pt}
\begin{tabular}{rlc}
\toprule
& Canonical instruction & Step codes \\
\midrule
0 & open the middle drawer of the cabinet & 30 30 30 27 30 30 \\
1 & open the top drawer and put the bowl inside & 3 3 3 27 3 3 \\
2 & push the plate to the front of the stove & 7 7 7 14 7 7 \\
3 & put the bowl on the plate & 13 13 13 27 13 13 \\
4 & put the bowl on the stove & 29 29 29 27 29 29 \\
5 & put the bowl on top of the cabinet & 16 16 16 27 16 16 \\
6 & put the cream cheese in the bowl & 5 5 5 9 5 5 \\
7 & put the wine bottle on the rack & 13 13 13 17 13 13 \\
8 & put the wine bottle on top of the cabinet & 16 16 16 17 16 16 \\
9 & turn on the stove & 12 12 12 27 12 12 \\
 
\bottomrule
\end{tabular}
\caption{\textbf{Canonical step codes} of the 10 \goal instructions (six slots of the meaning codebook). \emph{Generated.}}
\label{tab:s:canon}
\end{table}

\paragraph{Action-code dictionary.}
\Cref{tab:s:dict} describes all 32 first-level action codes by the \goal demonstration segments they label.
The codebook is learned without labels, and yet most codes have a consistent gripper state and direction of motion: code 27 is the only code that closes the gripper (in 91\% of its segments), codes 30 and 10 open it, and among the frequent codes, 25 reaches forward, 7 descends, 0 lifts and 18 carries the object forward and up with the gripper closed.
Codes are shared across tasks (half of them are used by 7 or more of the 10), which lets chains of different tasks share their approach and grasp.

\paragraph{MDL versus gripper-event segmentation.}
\label{sec:s:codebook}
Before adopting MDL segmentation we compared it with a codebook of the same size whose segments are cut at gripper open/close events, a common hand-designed choice.
\Cref{tab:s:codebook} measures both on held-out demonstrations.
The gripper-event codebook gives short chains (3 segments) that are more consistent within a task, so a held-out demonstration's task is easier to recognize from its chain (73\% vs.\ 62\%).
The MDL codebook halves the reconstruction error of the motion, and its chains reuse far more pieces across tasks (74\% vs.\ 35\% of 3-grams also occur in another task), which composition needs; without ever seeing gripper events, it places a boundary within 3 steps of 87\% of them.
Its granularity comes from the data: from 16-step windows the median segment converges to 13 steps within two rounds, and coarser initializations shrink in the same direction but more slowly (to 16 and 38 steps after 8 rounds).
Gripper-event chains are also easier for a planner to predict: on an earlier version of the model built on SmolVLA's language model, offline first-token accuracy was 0.80 with gripper-event codes and 0.59 with MDL codes, at similar actor validation loss.
We chose MDL because it needs no hand-designed event rule and gives finer, more reusable codes; the planners of the discrete-versus-continuous comparison reach 0.66 first-token accuracy on them (\paper{sec:exp:dev}).

\end{document}